# SANIP: Shopping Assistant and Navigation for the visually impaired

Monali Ahire
Electronics &Telecommmunication
Vishwakarma Institute of Technology Pune, India

*monali.ahire18@vit.edu*

Amey Chavan
Electronics &Telecommmunication
Vishwakarma Institute of Technology Pune, India

*amey.chavan18@vit.edu*

Favin Fernandes
Electronics &Telecommmunication
Vishwakarma Institute of Technology Pune, India

*favin.fernandes18@vit.edu*

Devashri Borse
Electronics &Telecommmunication
Vishwakarma Institute of Technology Pune, India

*devashri.borse18@vit.edu*

Shubham Deshmukh
Electronics &Telecommmunication
Vishwakarma Institute of Technology Pune, India

*shubhum.deshmukh18@vit.edu*

Prof Jyoti Madake
Electronics &Telecommmunication
Vishwakarma Institute of Technology Pune, India

*jyoti.madake@vit.edu*

***Abstract*— *The proposed shopping assistant model SANIP is going to help blind persons to detect hand held objects and also to get a video feedback of the information retrieved from the detected and recognized objects. The proposed model consists of three python models i.e. Custom Object Detection, Text Detection and Barcode detection. For object detection of the hand held object, we have created our own custom dataset that comprises daily goods such as Parle-G, Tide, and Lays. Other than that we have also collected images of Cart and Exit signs as it is essential for any person to use a cart and also notice the exit sign in case of emergency. For the other 2 models proposed the text and barcode information retrieved is converted from text to speech and relayed to the Blind person. The model was used to detect objects that were trained on and was successful in detecting and recognizing the desired output with a good accuracy and precision.***

**Keywords: Yolov4, PyTesseract, OCR, pyzbar, pyttx3**

## I. Introduction

According to the World Health Organization, “285 million people are estimated to be visually impaired worldwide”. Several technologies such as automatic text readers, Braille note makers, and navigation assistance canes have been developed to assist the visually impaired. Concurrent advances in computer vision and hardware technologies provide opportunities for a visual-assistance system that can be used in multiple contexts.

Lot of electronic products are introduced for the visually impaired but all have some sort of drawbacks such as complexity in operation, need of more practice, higher cost, expensive design methodology and installation, non-optimized data, more time consuming and tough maintenance. By considering these issues, if the assistant System is used for visually impaired and blind people, it will be really worthy. Shopping is one of the interesting things for every human. But this simple task cannot be easily achieved by the blind. They need others' help for satisfying their own requirements. Object detection and Text detection is the simplest and efficient technology which can be used for object detection and identification in many applications such as supply chain management, objects tracking, anti-theft applications, logistics, warehousing etc. This can be used effectively for blind at the time of shopping and greatly improves the lifestyle of them. The goal of every product is nothing but to detect and recognize it. So the proposed system is effectively designed by considering these aspects in mind. It can be easily implemented in the users mobile where all kinds of things are available under one roof.

For the betterment of the visually impaired we have been developing interfaces and algorithms to assist the visually impaired with a focus on Shopping. This paper describes the algorithms that we have incorporated into this assistant system so that it can be used in multiple contexts.

## II. Related Work

Andrew W. Williams implemented a barcode-based solution comprising a combination of off-the-shelf components, such as an Internet- and Bluetooth-enabled cell phone [2], text-to-speech software and a portable barcode reader.

S. Advani *et al*. "A Multitask Grocery Assist System [5] for the Visually Impaired: Smart glasses, gloves, and shopping carts provide auditory and tactile feedback. In this paper the authors proposed a Multitask Grocery Assistance System for the visually impaired using smart glasses, gloves, and shopping carts to provide auditory and tactile feedback.

Other applications of the assistive systems are product recognition, person detection and food recognition.

Ms Swathika R designed an android application which detects EXIT signs in a shopping mall, shops, etc.[1] in case of an emergency situation and guide blind people along a safe path through audio output Uses Image processing, edge detection, region detection, template matching, Maximally Stable External Regions (MSER), Optical Character Recognition (OCR) to detect the sign boards of ‘EXIT’.

Aqsa Aziz proposed a smart shopping facilitator [4] for blind using Raspberry PI3. The authors describe a system that helps the blind people to identify and purchase products using QR code and the audio instruction will assist them inside a supermarket.

P. A. Zientara *et al*., "Third Eye: A Shopping Assistant [3] for the Visually Impaired. In this illustrative scenario, cameras in smart glasses and a glove capture video that is streamed wirelessly to a server for video analysis. The system uses the results to guide the user through either audio commands based on images from the glasses to navigate toward the desired item’s proximity or vibrations in the glove to acquire the item.

## III. Dataset & Algorithms

### A) Dataset

For this project we have collected images of: -
Household – Parle-G, Lays, Tide 30 images each
Shopping – Shopping Cart 30 images
Signs – EXIT sign 30 images

A total of 150 images collected.
The dimensions of the images were resized to 1500x1500 pixels with horizontal and vertical resolution of 96 dpi.
The images were stored for the detection in the jpg format.

Since the object were going to be trained on a yolo architecture the images had to be annotated with bounding boxes for the architecture. For this we used an image labeling tool known as LabelImg.

LabelImg is a graphical image annotation tool. It is written in Python and uses Qt for its graphical interface. Annotations are saved as XML files in PASCAL VOC format, the format used by ImageNet. Besides, it also supports YOLO format.

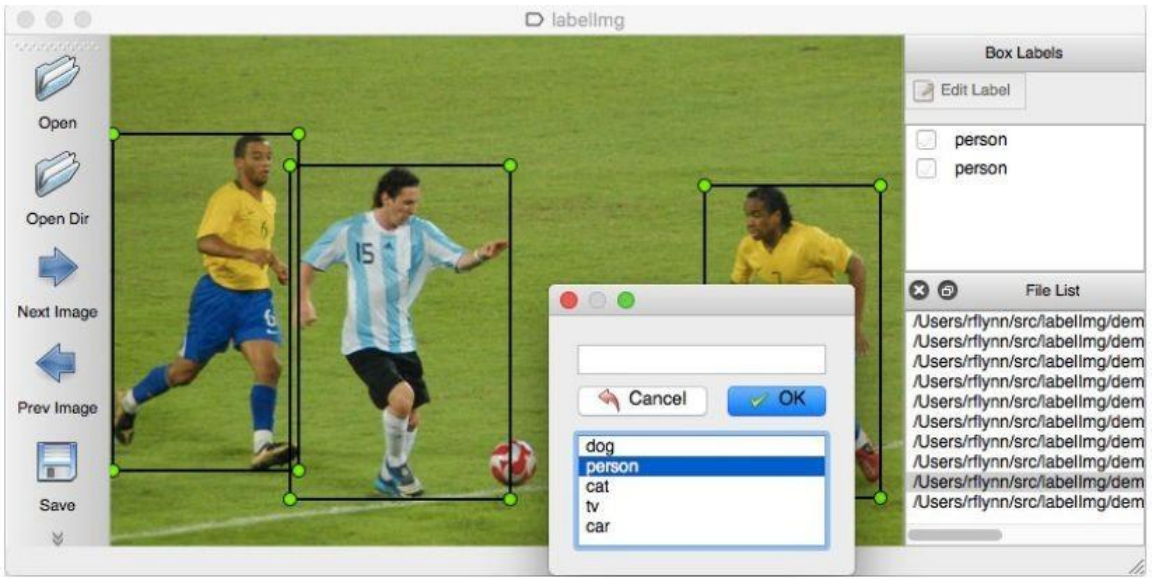

Fig .1. LabelImg Software

A txt file of YOLO format will be saved in the same folder as the image with the same name after the annotation was done. The text file will have the top and bottom X, Y coordinates of the bounding boxes. A file named "classes.txt" is also saved to that folder too. "classes.txt" defines the list of class names that your YOLO label refers to.

### B) Algorithms

The following algorithms were all implemented using Python language. The platform for it was Google Colab and Jupyter notebook.

1. Custom Object detection using YOLOv4 tiny

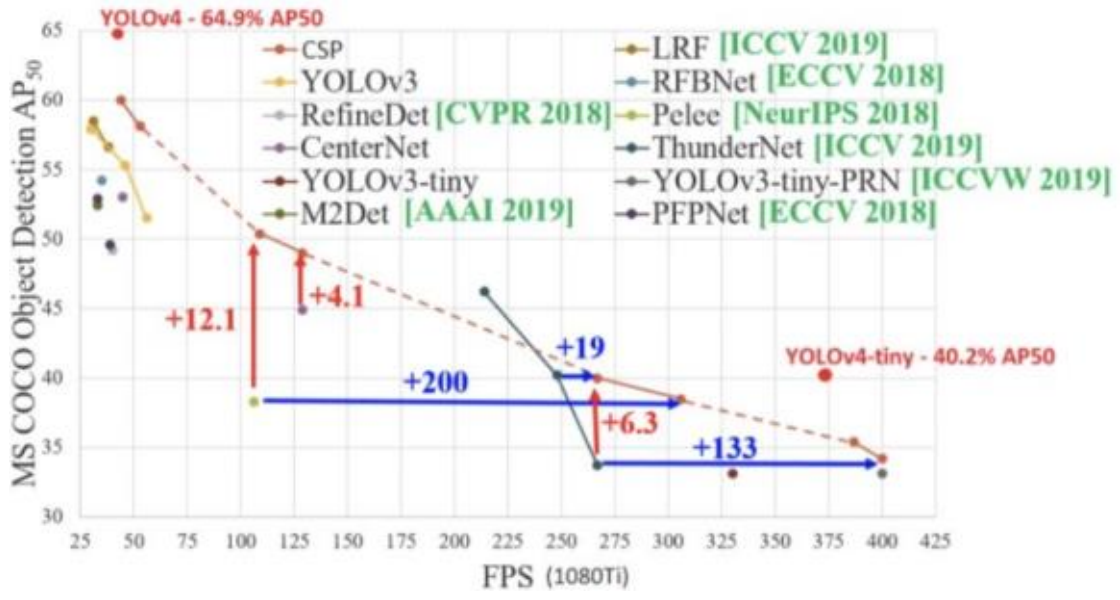

Fig .2. Different performance rates of various YOLO versions

The “You only look once v4” (YOLOv4) is one type of object detection method in deep learning. YOLOv4-tiny is proposed based on YOLOv4 to simplify the network structure and reduce parameters, which makes it suitable for developing on mobile and embedded devices. To improve the real-time of object detection, a fast object detection method is proposed based on YOLOv4-tiny. It firstly uses two ResBlock-D modules in the ResNet-D network instead of two CSPBlock modules in Yolov4- tiny, which reduces the computation complexity. Secondly, it designs an auxiliary residual network block to extract more feature information of objects to reduce detection error. In the design of auxiliary network, two consecutive 3x3 convolutions are used to obtain 5x5 receptive fields to extract global features, and channel attention and spatial attention are also used to extract more effective information. In the end, it merges the auxiliary network and backbone network to construct the whole network structure of improved YOLOv4- tiny. Simulation results show that the proposed method has faster object detection than YOLOv4-tiny and YOLOv3-tiny, and almost the same mean value of average precision as the YOLOv4-tiny. It is more suitable for real-time object detection, especially for developing on embedded devices.

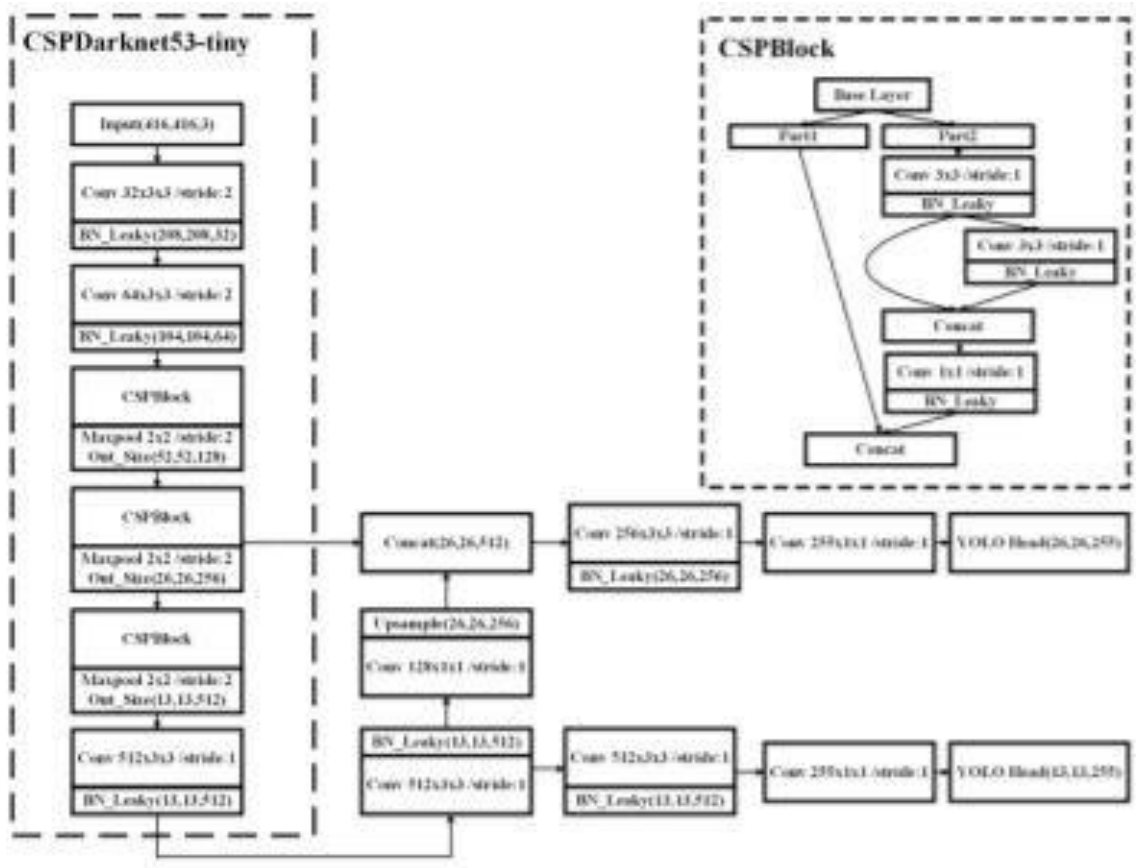

Fig .3.Architecture of YOLOv4 tiny

2. Text detection using Pytesseract

Python-tesseract is an optical character recognition (OCR) tool for python. That is, it will recognize and “read” the text embedded in images.

Python-tesseract is a wrapper for Google’s Tesseract-OCR Engine. It is also useful as a stand-alone invocation script to tesseract, as it can read all image types supported by the Pillow and Leptonica imaging libraries, including jpeg, png, gif, bmp, tiff, and others. Additionally, if used as a script, Python-tesseract will print the recognized text instead of writing it to a file.

**Functions**

- **get_languages** Returns all currently supported languages by Tesseract OCR.
- **get_tesseract_version** Returns the Tesseract version installed in the system.
- **image_to_string** Returns unmodified output as string from Tesseract OCR processing
- **image_to_boxes** Returns result containing recognized characters and their box boundaries
- **image_to_data** Returns result containing box boundaries, confidences, and other information. Requires Tesseract 3.05+. For more information, please check the Tesseract TSV documentation
- **image_to_osd** Returns result containing information about orientation and script detection.
- **image_to_alto_xml** Returns result in the form of Tesseract’s ALTO XML format.
- **run_and_get_output** Returns the raw output from Tesseract OCR. Gives a bit more control over the parameters that are sent to tesseract.

3. Barcode/ QR code detection using Pyzbar

Pyzbar is an OpenCV module that is used to decode the Barcode/ QR code and it is distributed under MIT license. Using this library, it illustrates type: whether it is barcode or QRcode, data: the alphanumeric information is hidden in QRcode, and location: edges of the Barcode/ QR in the image.

4. Python text to speech for voice feedback

Pyttsx3 is a text-to-speech conversion library in Python. Unlike alternative libraries, it works offline, and is compatible with both Python 2 and 3.

pyttsx3 includes drivers for the following text-to-speech synthesizers. Only operating systems on which a driver is tested and known to work are listed. The drivers may work on other systems.

- SAPI5 on Windows XP and Windows Vista and Windows 8,8.1 , 10
- NSSpeechSynthesizer on Mac OS X 10.5 (Leopard) and 10.6 (Snow Leopard)
- espeak on Ubuntu Desktop Edition 8.10 (Intrepid), 9.04 (Jaunty), and 9.10 (Karmic)

The **pyttsx3.init()** factory functions to get a reference to a **pyttsx3.Engine** instance. During construction, the engine initializes a **pyttsx3.driver.DriverProxy** object responsible for loading a speech engine driver implementation from the **pyttsx3.drivers** module. After construction, an application uses the engine object to register and unregister event callbacks; produce and stop speech; get and set speech engine properties; and start and stop event loops.

# IV. Methodology

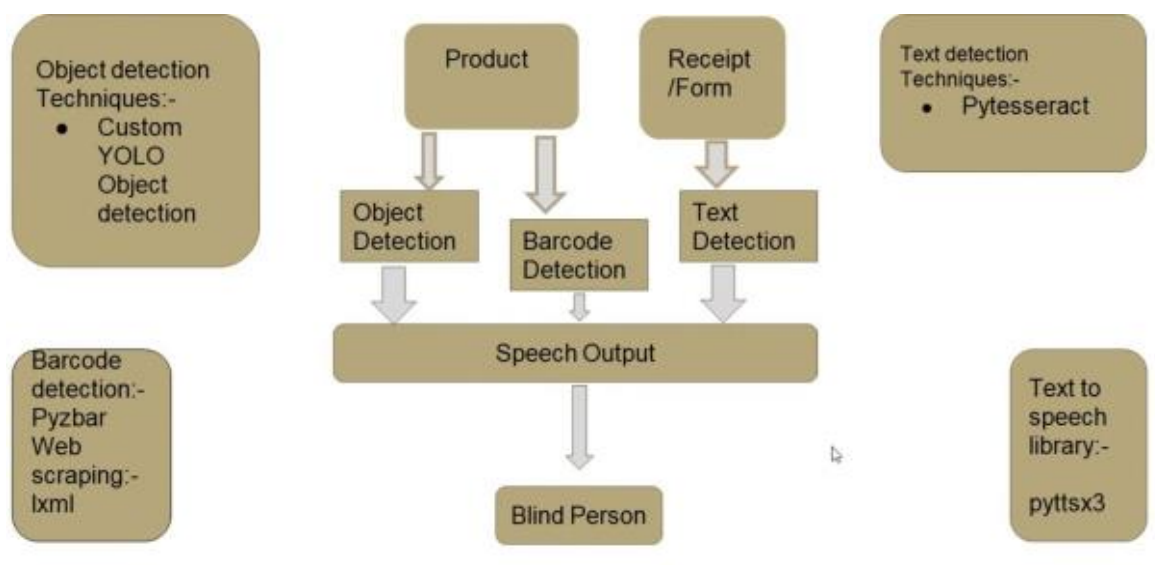

Fig .4. Proposed Methodology

1. Object Detection

The custom dataset consisting of Lays, Parle-G, Tide, Exit, sign and shopping Cart with their annotated text file are trained on the darknet framework.
The Training of the model was done on Google Colab. The free GPU provided for the training was Tesla T4.

In order to set up our Darknet environment the following dependencies were installed:

- OpenCV
- Cuda Toolkit
- GPU resources
- cuDNN

For the Yolo model to train on the darkent framework Yolov4-tiny custom configuration file was used. In the Yolov4-tiny custom config file the following changes were made:

- num_classes = 5
- max_batches = 5 x 2000= 10000
- iteration steps = (0.8 x 10000),(0.9 x 10000)
- layer filters = (num classes + 5)*3

The training weights used along with the configuration file is yolov4tiny.conv.29 which has the initial features of products like edges, RGB etc.

The model was trained in 7200 batches and the 3000 batches weights was considered the best for the object detection Model.

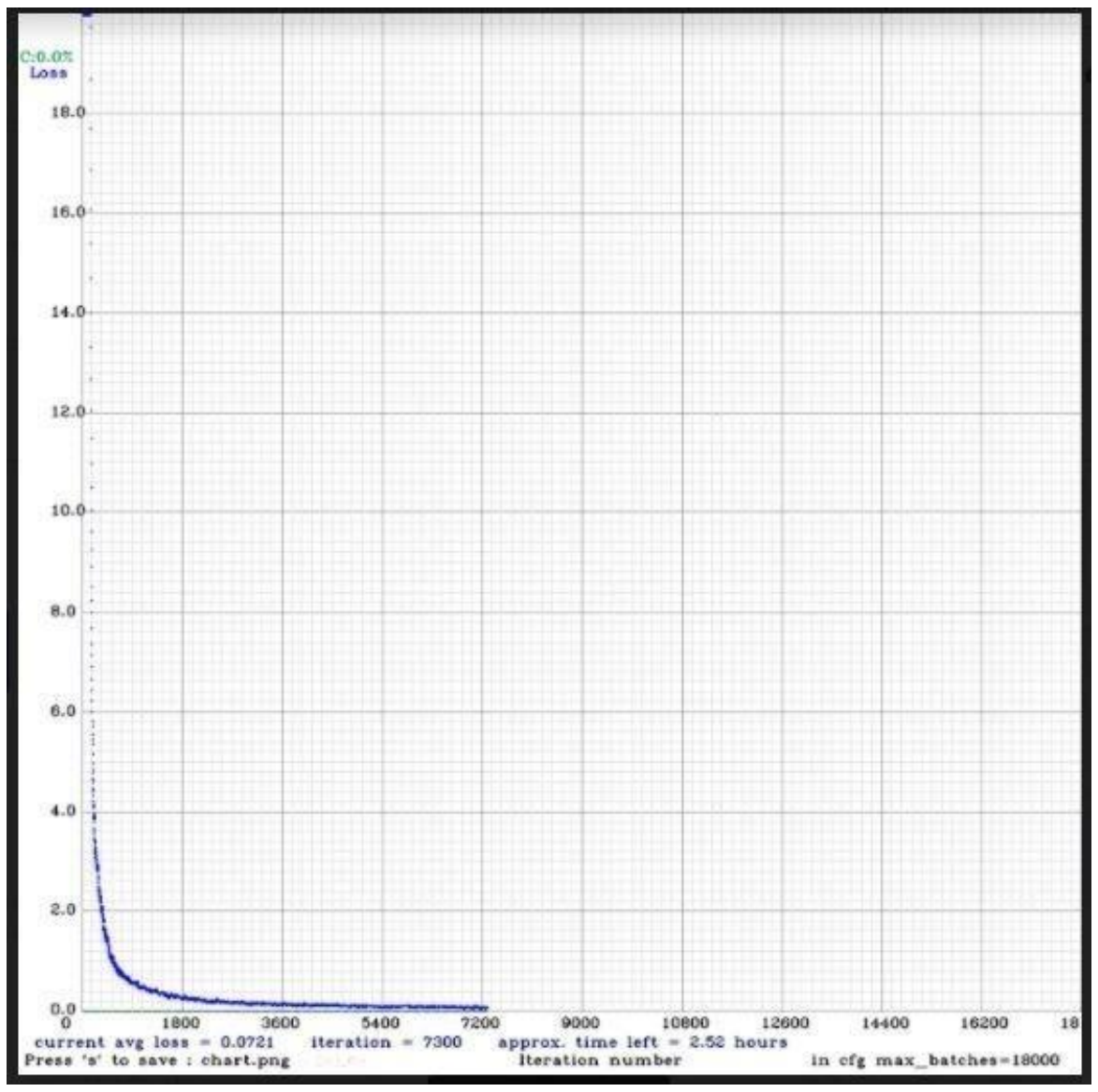

Fig .5 Loss vs Batches graph

The object detection code was written in python using OpenCV libraries. The weights were tested using the webcam.

The model successfully detected the product in real time.

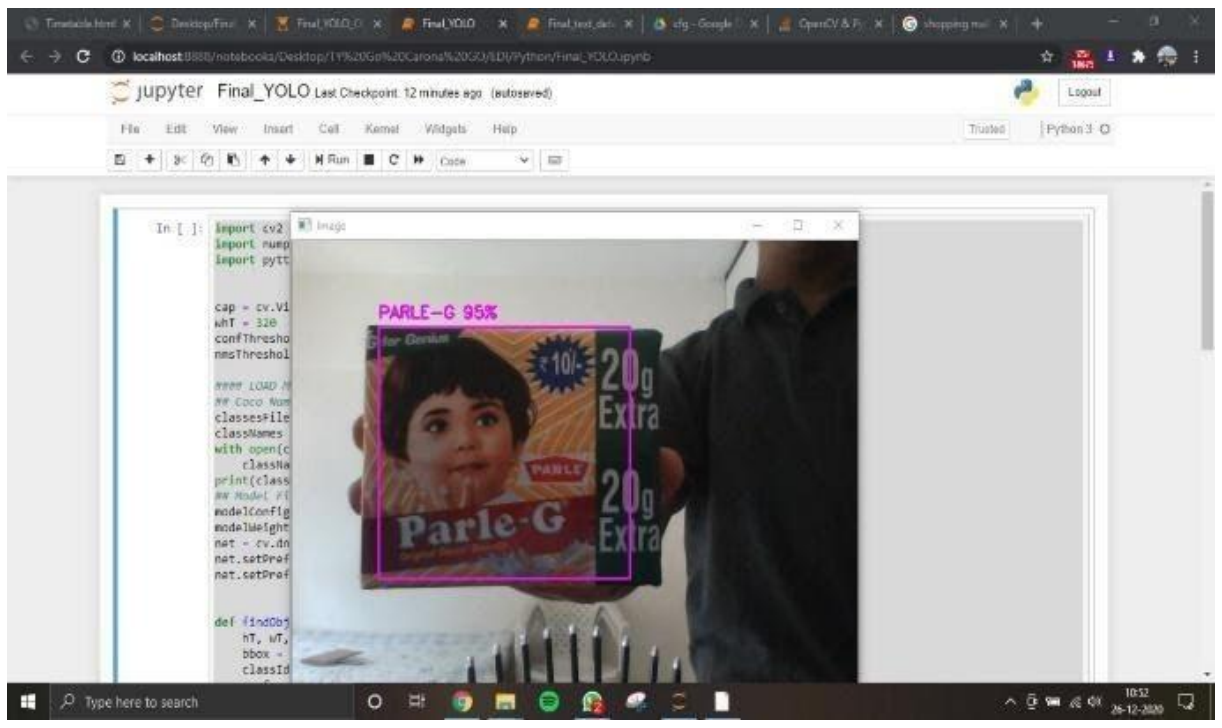

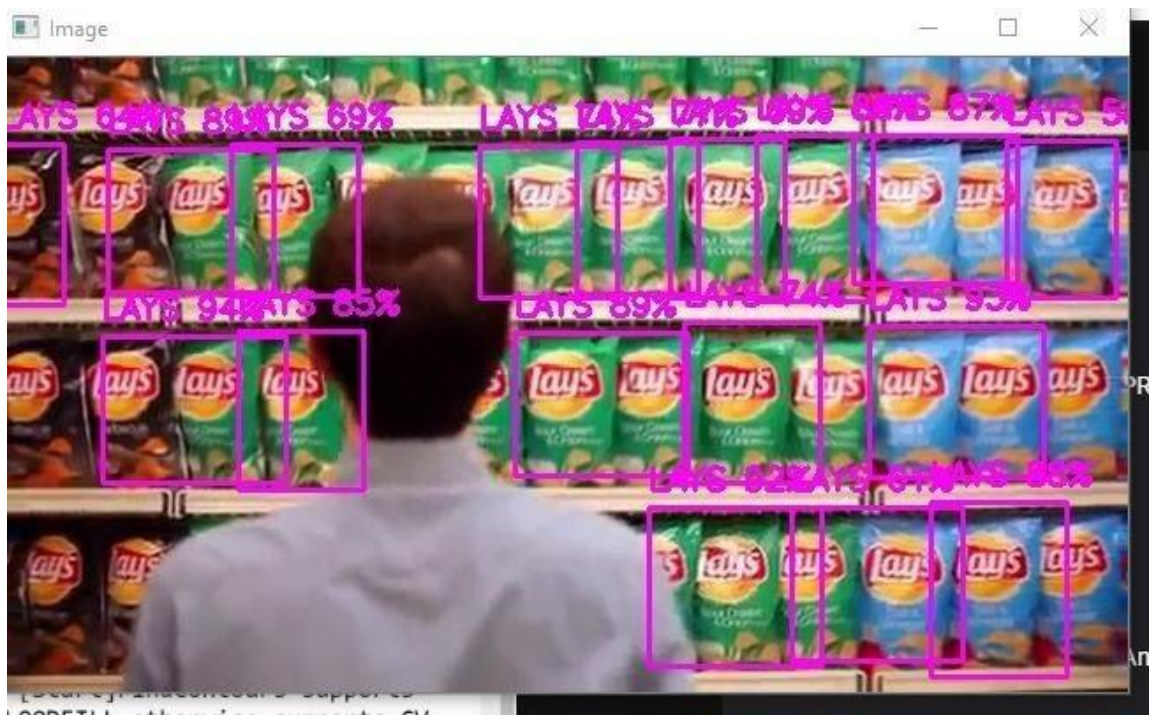

Fig .6 Sanip app Object detection results

After Successful detection, the recognized object is given as a speech output using the pyttsx3 library for voice feedback

## 2. Text Detection

The code for the text detection model was written in python and tested on Jupyter notebook. Using the pytesseract library we could input an image of receipts and detect strings of text

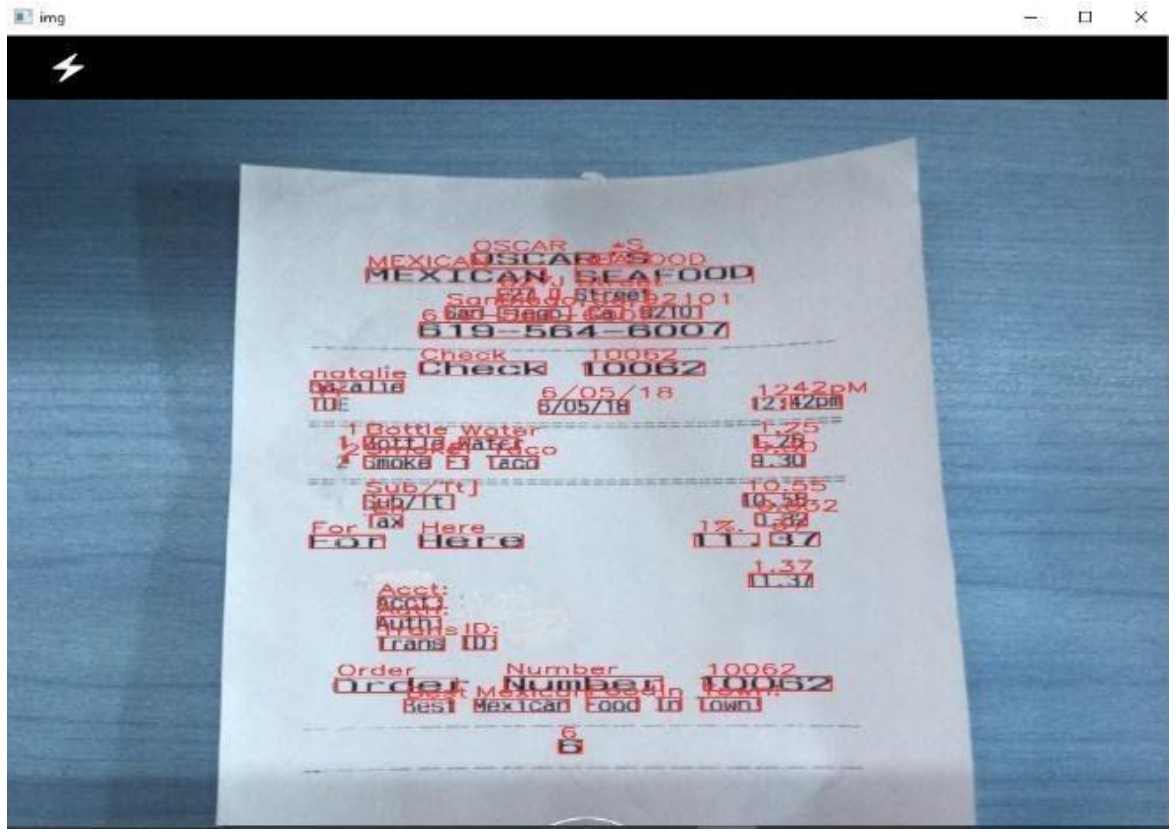

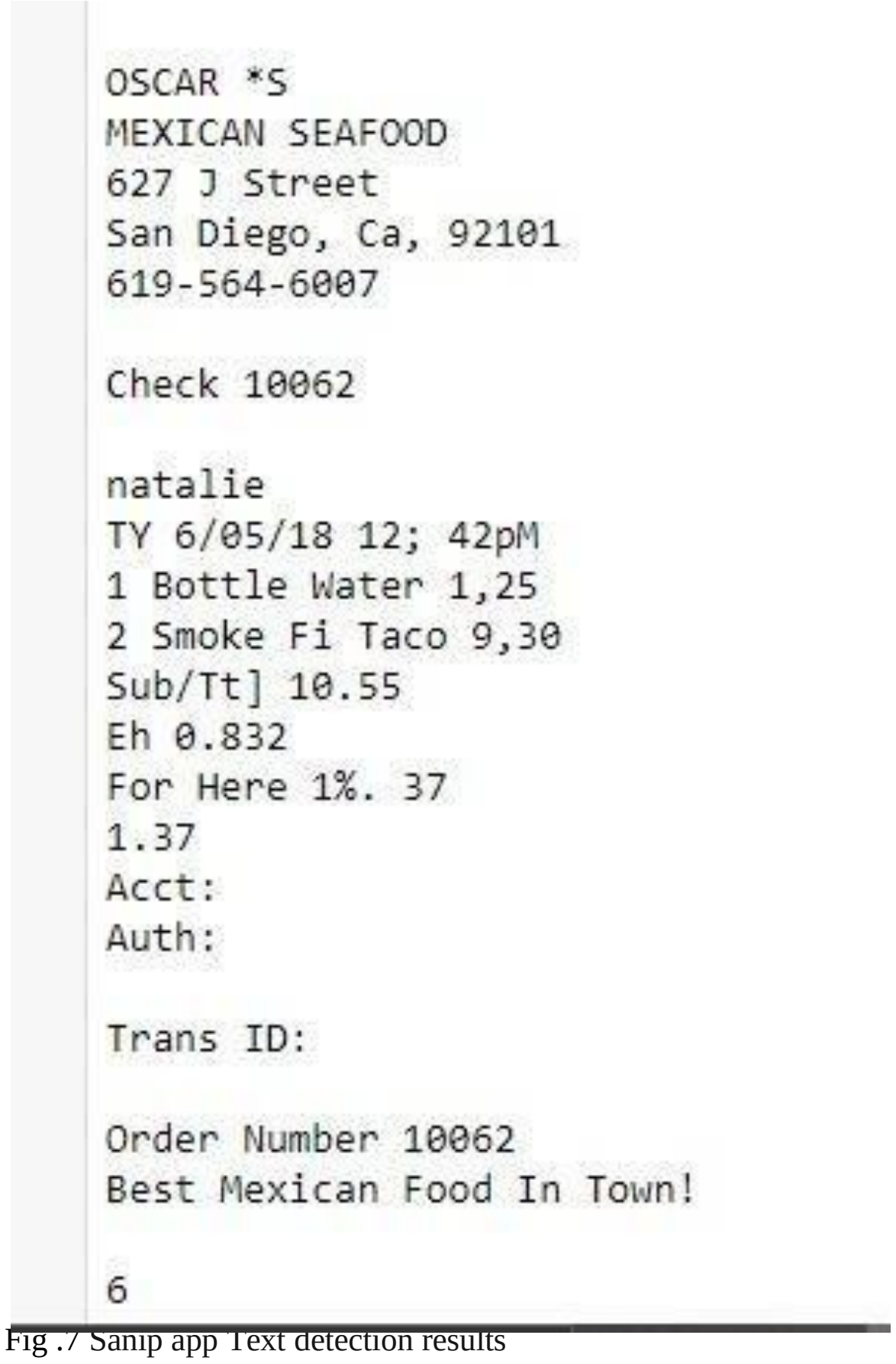

Fig .7 Sanip app Text detection results

After Successful detection and conversion, the text generated is then converted from text to speech using the pyttsx3 library for voice feedback.

### 3. Barcode/QR code Detection

The code for the following model was written in Jupyter Notebook using Python and it was tested in real time using a webcam. The output of the barcode/QR code detected with its information is then converted from text to speech using pyttsx3.

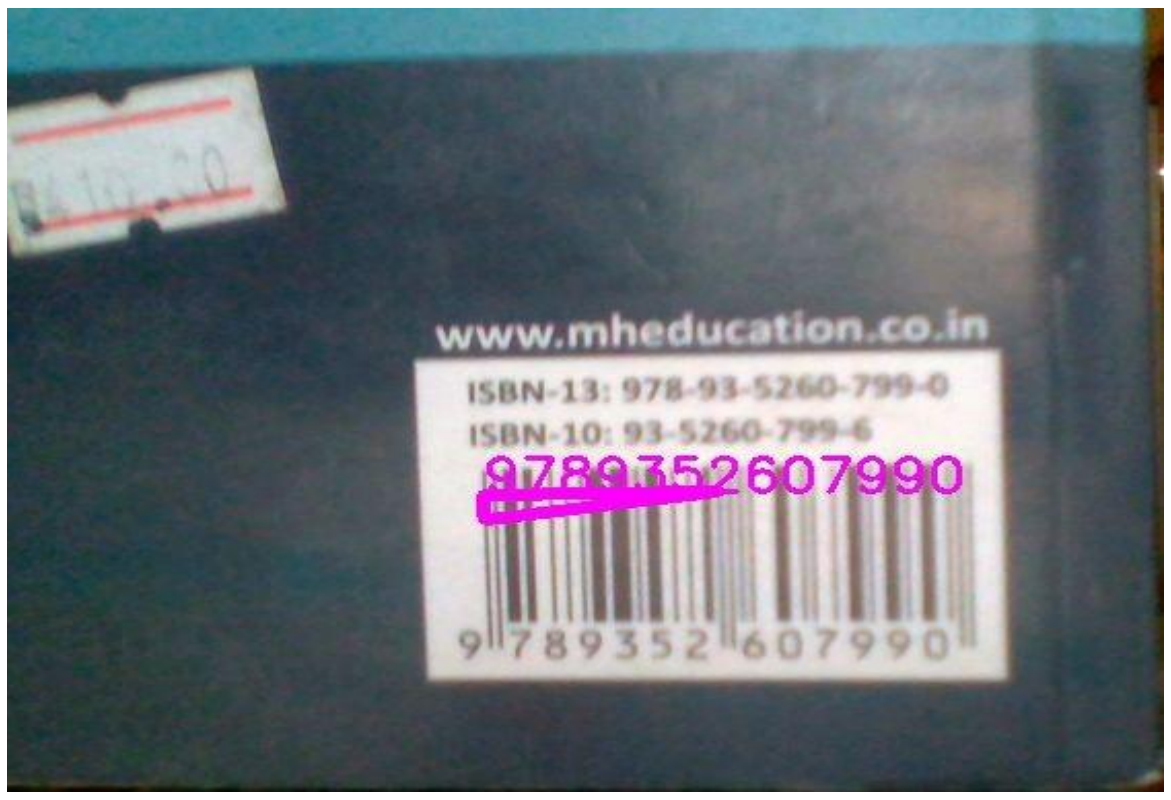

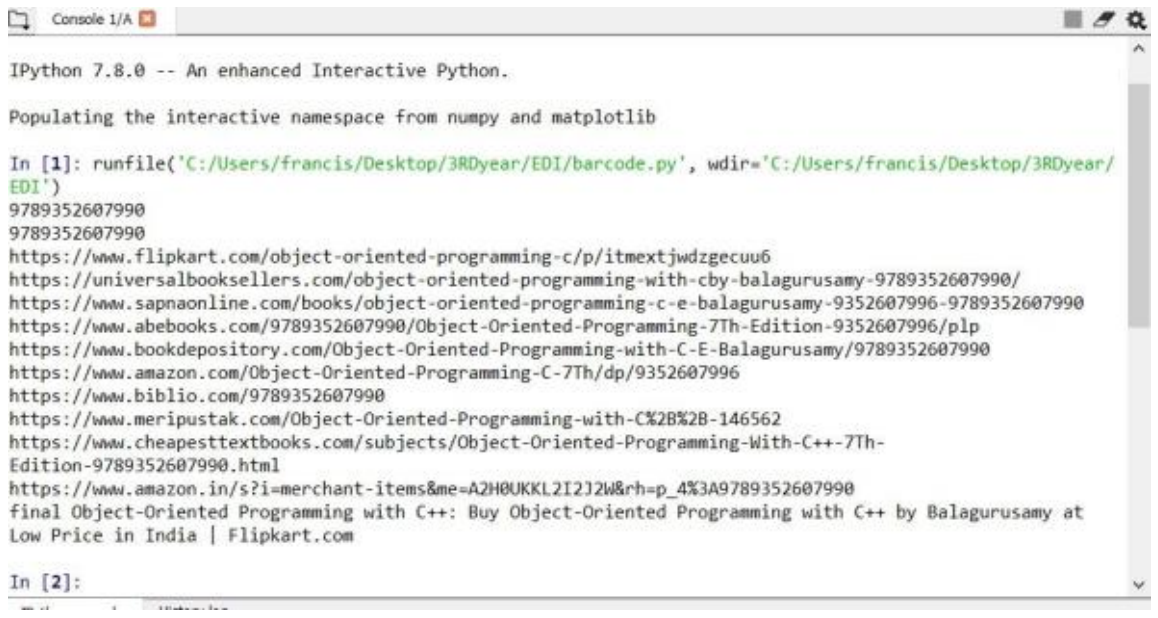

Fig .8 Sanip app Barcode detection results

## V. Conclusion

In this project we had proposed a project comprising of 3 detection models i.e. Object, text and Barcode/QR code. We have successfully detected and recognized the objects and texts along with the barcode and QR code.
The speech output also was relayed without any distortion as pyttsx3 is able to work offline

## VI. Future Scope

For the future scope we'll be implementing the 3 proposed models onto an app using tensorflow.
along with it we will be gathering more images of different classes to train on to improve accuracy and also increase the detection of objects.